\documentclass[letterpaper, 10 pt, journal, twoside]{IEEEtran}

\IEEEoverridecommandlockouts                              % This command is only needed if
\usepackage{tabularx}
\usepackage{amsthm}
\usepackage{booktabs}
\usepackage{multirow}
\usepackage{bbm}
\usepackage[dvipsnames]{xcolor}

\usepackage{subcaption} % apparently overrides IEEE captions
\usepackage{enumerate}
\usepackage{mathrsfs}

\usepackage[bottom=57pt,top=54pt, left=54pt, right=54pt]{geometry}   %57-43
\usepackage{graphicx}
\usepackage{adjustbox}
\usepackage{pbox}
\usepackage{dblfloatfix}
\usepackage{graphicx}
\usepackage{hyperref}
\usepackage{caption}
\usepackage{fancyhdr}
\fancypagestyle{arxivfooter}{
  \fancyhf{} % clear header/footer

  \fancyfoot[L]{\footnotesize
    This work has been accepted for publication in IEEE Robotics and Automation Letters. 
   \textcopyright\ IEEE. The final version is available at IEEE Xplore: \url{https://doi.org/DOI_HERE}
 }
}

\author{Andreea Tulbure, René Zurbrügg, Timm Grigat, Marco Hutter % <-this % stops a space
\thanks{
This research was supported by the ETH AI Center and by ETH Zurich Research Grant No. 22-2 ETH-47.} %Use only for final RAL version
\thanks{All authors are with the Robotics System Lab, ETH Zürich, Switzerland.
        {\tt\footnotesize Email: atulbure@ethz.ch}}%
\thanks{Digital Object Identifier (DOI): see top of this page.}
}

\usepackage{xcolor}
\usepackage{graphicx}
\usepackage{hyperref}
\usepackage{amsmath}
\usepackage{cleveref}
\usepackage{amssymb}
\usepackage{bm}
\usepackage{mathtools}
\usepackage{epsfig}
\usepackage{epstopdf}
\usepackage{todonotes}
\usepackage{url}
\usepackage{cite}
\usepackage{multirow}
\usepackage{float}

\usepackage{amssymb}% http://ctan.org/pkg/amssymb
\usepackage{pifont}% http://ctan.org/pkg/pifont
\usepackage{soul}

\newcommand{\removed}[1]{}
\title{\LARGE \bf
LLM-Handover: Exploiting LLMs for Task-Oriented Robot-Human Handovers
}

\begin{document}

\maketitle
%\begin{center}
%    \vspace{-0.5cm}
%    \fbox{\parbox{0.95\linewidth}{
%        \footnotesize
%        This work has been accepted for publication in IEEE Robotics and Automation Letters. 
%        \textcopyright\ IEEE. The final version is available at IEEE Xplore: \url{https://doi.org/DOI_HERE}
%    }}
%\end{center}
%%%%%%%%%%%%%%%%%%%%%%%%%%%%%%%%%%%%%%%%%%%%%%%%%%%%%%%%%%%%%%%%%%%%%%%%%%%%%%%%
\begin{abstract}
Effective human-robot collaboration depends on task-oriented handovers, where robots present objects in ways that support the partner’s intended use. However, many existing approaches neglect the human’s post-handover action, relying on assumptions that limit generalizability.
To address this gap, we propose LLM-Handover, a novel framework that integrates large language model (LLM)-based reasoning with part segmentation to enable context-aware grasp selection and execution. Given an RGB-D image and a task description, our system infers relevant object parts and selects grasps that optimize post-handover usability. 
To support evaluation, we introduce a new dataset of 60 household objects spanning 12 categories, each annotated with detailed part labels.
We first demonstrate that our approach improves the performance of the used state-of-the-art part segmentation method, in the context of robot-human handovers.
Next, we show that LLM-Handover achieves higher grasp success rates and adapts better to post-handover task constraints. During hardware experiments, we achieve a success rate of $83\%$ in a zero-shot setting over conventional and unconventional post-handover tasks. Finally, our user study underlines that our method enables more intuitive, context-aware handovers, with participants preferring it in $86\%$ of cases.
\end{abstract}


\section{Introduction}\label{sec:Introduction}

\IEEEPARstart{A}s robots become more common in everyday settings, their ability to collaborate with humans on joint tasks becomes increasingly important.
Recent research in human-robot interaction explores these challenges, with object handovers being a key component for successful collaboration \cite{trendsCastro2021}. 
These handovers form the basis for many joint activities that require both physical coordination and contextual understanding \cite{Ortenzisurvey}. For instance, observations of human-to-human handovers reveal that people often anticipate each other's intended use of an object by interpreting the surrounding context \cite{Strabala2013seamless, Ortenzisurvey}. 
This ability, known as task-orientation, becomes especially critical in environments like factories, surgeries, or construction sites, where one partner may have limited mobility or freedom to adjust their actions.

\begin{figure}
    \vspace{-5pt}
    \centering
     \includegraphics[width=0.9\linewidth]{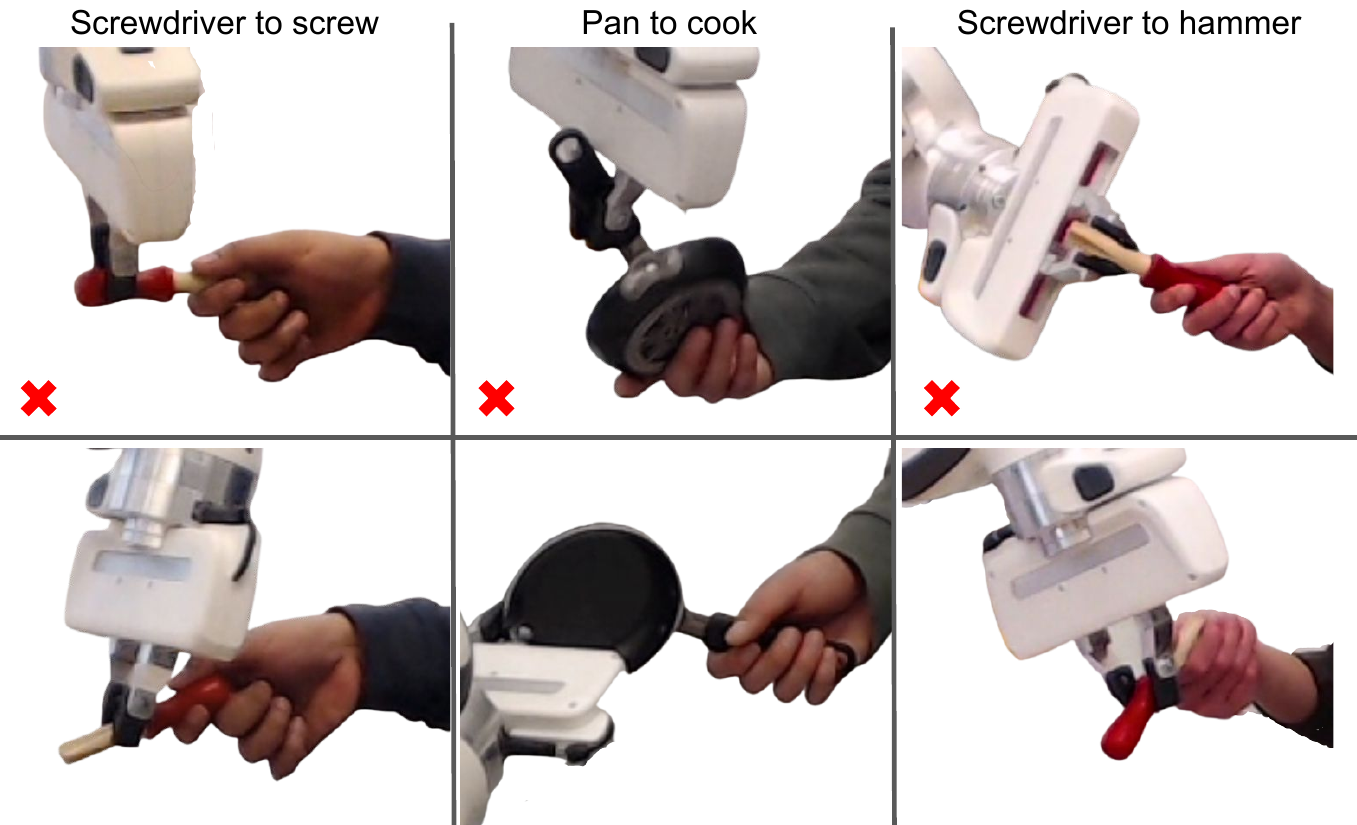}
    \vspace{-4pt}
    \caption{Examples of task-oriented (bottom) vs. non-task-oriented (top) robot-human handovers for conventional (left, center) and unconventional (right) object-task pairs.}
    \label{fig:handover}
\vspace{-20pt}
\end{figure}

Therefore, the challenge of robot-human handovers goes beyond simply transferring an object \cite{Meiying2022}. 
The aim is to facilitate task-oriented handovers, where the object is delivered in a manner that allows the human to use it immediately in the subsequent task, minimizing the need for in-hand or bimanual adjustments \cite{ortenzi2020grasp}. As shown in Fig.~\ref{fig:handover}, the top row presents tools handed over in orientations that hinder immediate use, while the bottom row depicts tools presented in a task-oriented manner.
To achieve this, the robot must account for task-specific details, which are closely related to the intended use of the object \cite{ortenzi2020grasp}.
However, in robot-human handovers, the post-handover task of the human is rarely accounted for \cite{Ortenzisurvey, Meiying2022, ortenzi2020grasp}. Most prior works in this direction infer human contact regions based on grasp affordances \cite{Meiying2022,chan2020affordance,contactHandover,ortenzi2020grasp, Aleotti2014, Ardon2020, Meng2022}. 
While most methods select grasps outside the human contact map under given assumptions, their generalizability is usually limited.
%%%%%%%%%%%%%%%%%%%%%%%%%%%%%%%
%%%%%%%%%%%%%%%%%%%%%%%%%%%%%%%
Recently, impressive zero-shot generalizability has been shown in task-oriented robot grasping, where the problem is to find the best robot grasp for a specific task. Those works exploit the semantic understanding of LLMs or Vision Language Models (VLMs) \cite{tang2025foundationgrasp,tang2023graspgpt,2023LanGrasp}. Although some of these methods \cite{tang2025foundationgrasp,tang2023graspgpt} show their suitability for robot-human handovers, they do not explicitly incorporate the task the human wants to perform, limiting their ability to perform reliable task-oriented handovers across a variety of objects. 

To address zero-shot generalization while modeling human intent in task-oriented handovers, we present \emph{LLM-Handover}—a framework that leverages LLMs to infer task-specific context information and select the most suitable robot grasps from a set of candidates. 
To inform grasp selection, we rely on part segmentation to extract spatial and semantic information.
However, current part segmentation networks fail to deliver reliable results in robotic scenarios, typically over-/under-segmenting parts.
To tackle this, we propose using an LLM-based reasoning pipeline on top of part segmentation. Overall, our contributions are:
\begin{enumerate}
    \item A novel task-oriented handover framework combining part segmentation with LLM-based reasoning
    \item An LLM-based reasoning pipeline that enhances existing part segmentation algorithms for robotic handovers.
    \item An extensive user study showing that incorporating the human’s post-handover task for LLM-based grasp selection significantly enhances perceived task understanding, user preference, and reduces regrasp frequency.
    \item An open-source RGB-D dataset with hand-labeled part masks for 60 objects across 12 categories\footnote{Available at \textit{https://andreeatulbure.github.io/llm-handover/}}.
\end{enumerate}

%%%%%%%%%%%%%%%%%%%%%%%%%%%%%%%
\begin{figure*}[ht!]
    \vspace{2pt}
    \centering    \includegraphics[trim=20 10 20 5, clip, width=0.98\linewidth]{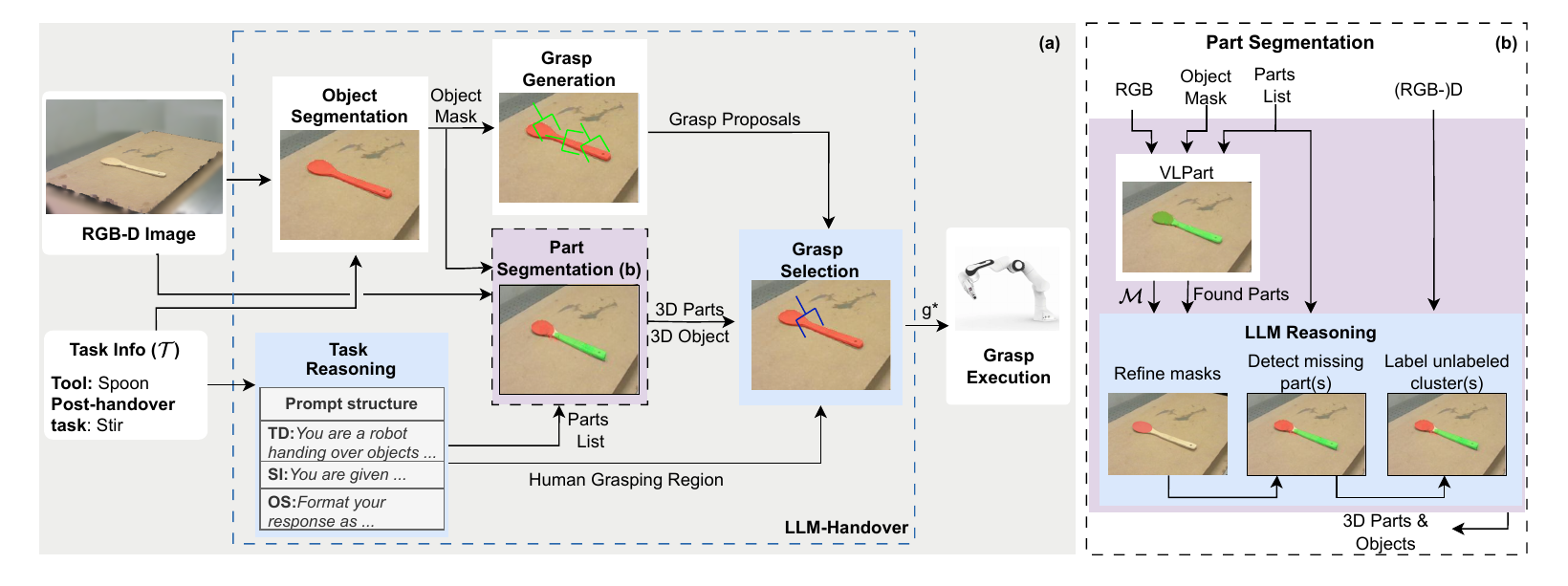}
    \vspace{-6pt}
    \caption{
    \textbf{Overview of LLM-Handover}.(a) High-level schematic with new components marked blue, existing white  and modified violet.
    Given a task ($\mathcal{T}$), \emph{Task Reasoning} uses an LLM with a prompt structured into Task Description (TD), Supporting Information (SI) and Output Structure (OS) to infer a post-handover task description, relevant parts and human and robot grasp regions. The relevant parts and human grasp region are streamed to subsequent modules.
    Given the RGB-D image, \emph{Object Segmentation} identifies the target object mask, and \emph{Grasp Generation} produces diverse candidate grasps. 
    \emph{Part Segmentation} combines the object mask, relevant parts and RGB-D image to generate 3D representations of both parts and object. These, together with the grasp candidates and the specified human grasp region, are fed into \emph{Grasp Selection}, which calls an LLM to pick the single best grasp for the task, which is executed.(b) Detailed view of \emph{Part Segmentation}, combining an existing segmentation algorithm, VLPart, with our proposed LLM-reasoning pipeline to refine the outputs.}
\label{fig:frameworkDiagram}
\vspace{-15pt}
\end{figure*}
%%%%%%%%%%%%%%%%%%%%%%%%%%%%%%%

\section{Related Work}

\subsection{Task-oriented Handovers}
Most prior works rely on the concept of grasp affordances for robot-human handovers \cite{Meiying2022,chan2020affordance,contactHandover,ortenzi2020grasp,Aleotti2014, Ardon2020,Meng2022}. Aleotti et al. \cite{Aleotti2014} manually select the object part that the human will grasp, while Ortenzi et al. \cite{ortenzi2020grasp} manually select the best robot grasp. 
Demonstrating better generalization capabilities and less reliance on human labels, Ardon et al. \cite{Ardon2020} optimize over affordances, task to perform, and mobility constraints of the human receiver. They cluster objects by shape, showing that their method generalizes to previously unseen but semantically similar objects. 
Furthermore, Meng et al. \cite{Meng2022} use a neural network to predict human grasp affordances, but ignore the post-handover task. They rely on heuristics to place the robot’s grasp opposite to the human interaction.

To go beyond heuristics, Wang et al. \cite{contactHandover} propose a neural network that predicts 3D human contact affordance maps, considering only object geometry, without task context. Robot grasps generated by a grasp planner are reranked to avoid human contact regions, with the highest-ranked grasp selected for execution. However, their evaluation is limited to simulation, and generalization is constrained by the size and diversity of the training dataset. Meiying et al. \cite{Meiying2022} train a neural network with tool-use demonstrations to capture task context. This allows the robot to adapt handover configurations across typical and unconventional scenarios. While some generalization is shown, the effectiveness on entirely different objects remains unclear.

\subsection{Task-oriented robot grasping}
While task-oriented robotic handovers explicitly consider human-centered factors, such as the human's intended task, task-oriented robotic grasping typically focuses on optimizing grasps for autonomous manipulation. Nonetheless, recent work has shown that robots can achieve high success rates in handover tasks and demonstrate strong zero-shot generalization across diverse objects \cite{tang2025foundationgrasp,tang2023graspgpt,2023LanGrasp}. These works employ LLMs \cite{tang2023graspgpt} or both LLMs and VLMs \cite{tang2025foundationgrasp,2023LanGrasp} to gather task-specific information.  

Tang et al.\cite{tang2023graspgpt} introduce a framework that prompts an LLM to produce natural language descriptions for novel object-task pairs.
These descriptions are leveraged to relate new concepts to known ones, enabling the robot to transfer learned grasp strategies to new object-task pairs. While in \cite{tang2023graspgpt}, just semantic task and object descriptions are used for grasp reasoning, in follow-up work \cite{tang2025foundationgrasp}, geometric information is employed as well. The LLM provides a semantic and geometric description of object-task pairs, and a VLM extracts features from images and the geometric descriptions to provide both textual reasoning and visual understanding of objects and affordances. A transformer-based evaluator with semantic and geometric branches is implemented to assess grasp candidates.
Authors demonstrate that their robot can execute successfully a "handover" task \cite{tang2025foundationgrasp,tang2023graspgpt}, grasping a pan at the body to leave space for the human to grasp the handle. 
Most similar to our work is \cite{2023LanGrasp}, where part segmentation VLMs \cite{owlvit} and the semantic understanding of LLMs are combined. The LLM determines the most suitable object part for a specific task, while the VLM segments the part from the RGB image. Even though the authors do not explicitly consider handover tasks, we think their framework can handle such tasks. However, since human-centered factors are not explicitly considered, these methods do not hand over objects reliably in different scenarios.

\section{Methodology} 
\label{sec:Method}

\subsection{Problem Formulation and Notation}
We consider a robot equipped with a two-finger parallel-jaw gripper operating in a tabletop environment. The input consists of an RGB-D image of a single object, and a natural language description $\mathcal{T}$ specifying the tool and its intended post-handover use. We assume the object is isolated in a clutter-free environment.
The goal is to predict a single task-compliant grasp pose $g^* \in SE(3)$ for the robot that enables effective handover. As part of the reasoning process, the framework produces part segmentation masks $\mathcal{M} \in \{1, \dots, P\}^{H \times W}$ that identify functionally relevant object parts. While not the primary goal, segmentation acts as an intermediate representation to support grasp selection conditioned on task semantics and object geometry.

\subsection{LLM-Handover}
A schematic of the proposed framework is depicted in Fig.~\ref{fig:frameworkDiagram}. 
The system is initialized with a natural language command ($\mathcal{T}$) specifying the tool and its intended post-handover use, and an RGB-D image.
As shown in Fig.~\ref{fig:frameworkDiagram}, Task Reasoning takes the command  ($\mathcal{T}$) and queries an LLM to generate a post-handover task description and a list of task-relevant functional object parts, and determine suitable grasp regions for both human and robot. The list of relevant parts and the human grasp region are streamed to other modules, while the other outputs are queried solely to facilitate chain-of-thought type reasoning. 
Simultaneously, the Object Segmentation generates a segmentation mask of the target object, which serves as input for Part Segmentation and Grasp Generation.
The Part Segmentation is a two-stage module. First, it runs an off-the-shelf part segmentation algorithm. Second, an LLM reasons about the obtained segmentation masks. 
As input, Part Segmentation gets the RGB-D image, the object mask, and the list of functional parts and outputs the 3D geometric information of the object and its constituent parts.
Finally, Grasp Generation produces a set of grasp candidates based on the object geometry. These proposals, along with the human grasp region from Task Reasoning and the 3D part and object information from Part Segmentation, are passed to the Grasp Selection module. This prompts an LLM to select the most appropriate grasp, which is then executed by the robot.

We detail Part Segmentation and Grasp Selection below. For Object Segmentation and Grasp Generation, we use existing open-source frameworks, namely LangSAM \cite{LangSAM} and EdgeGrasp \cite{huang2022edge}. The Grasp Execution module contains the motion planner and low-level controller.

\subsubsection{Part Segmentation}\label{sec:part_seg}
This module provides 3D spatial information about the object and its parts to support grasp selection. As shown in Fig.~\ref{fig:frameworkDiagram}(b), the algorithm consists of two stages: (1) A standard part segmentation algorithm to produce candidate part masks, and (2) prediction refinement via LLM reasoning using spatial information.
While we used VLPart~\cite{peize2023vlpart} to generate part masks, the framework is compatible with any segmentation algorithm.

First, we crop the RGB image around the detected object mask to focus the part segmentation on the relevant region. VLPart processes the cropped image to generate initial 2D part masks, which—along with their part names and corresponding depth data—are passed to the LLM for spatial reasoning. To facilitate spatial understanding, the 2D masks are fused with the depth data to compute 3D geometric properties of both the object and its parts.
The LLM-reasoning consists of three stages: (1) \emph{refinement} of predicted masks, (2) \emph{detection} of missing parts, and (3) \emph{labeling} of unlabeled regions. Each step is described below.

\emph{Refine part masks:}
\label{sec:refine_masks}
In this step, we verify whether each predicted mask satisfies two criteria: (i) it remains within the original object mask, with a small tolerance, and (ii) it does not exhibit semantic contradictions. For example, in the case of a mug, significant overlap between the rim and body is acceptable, whereas for a pan, such overlap between the cooking surface and the handle is not. 
While the first criterion is straightforward to check, the second requires semantic reasoning, which we delegate to the LLM. If a contradiction is identified, the LLM assesses which part the given mask is more likely to represent. This two-step refinement process yields updated, consistent part masks.

\emph{Detect missing parts:}
\label{sec:missing-parts}
After refining the initial part masks, we check for missing parts, which can result from ambiguity in the functional part list or segmentation errors, particularly for geometrically similar parts (such as the shaft and tip of a screwdriver).
If parts are missing, we subtract all labeled points from the object point cloud and apply DBSCAN to find remaining clusters. If no clusters remain, the missing part is left unidentified. If only one unlabeled cluster is found and the missing parts cannot be reliably distinguished based on their spatial extent or geometric center, we assign a merged label (e.g., "shaft+tip") to the cluster. 
If cluster-part matching is ambiguous or multiple clusters exist, we query an LLM to infer part-cluster assignments based on spatial layout. This step is computationally more expensive, but necessary when simpler heuristics are insufficient.

\emph{Label unlabeled cluster:}
\label{sec:unlabelled}
After handling missing parts, we check whether any significant portion of the object remains unlabeled. If an unlabeled cluster exists, we query the LLM to determine whether it belongs to an existing or a new, previously unrecognized part. If the first is the case, the corresponding part mask is updated with the unlabeled region; otherwise, the new label is added to the part set.

\subsubsection{Grasp Selection}
\label{sec:grasp-selection} 
The Grasp Selection module takes as input (i) the 3D geometric information of the object and its parts, (ii) the part for human grasping as identified by Task Reasoning, and (iii) a set of feasible grasps generated in Grasp Generation. To reduce computational load, we first apply furthest point sampling to select the $k$ most spatially diverse grasps. We then verify whether at least two of these grasps are separated by a minimum threshold, which we empirically define as one-third of the dominant side length of the object. If this condition is not met, this module lacks sufficient spatial diversity for meaningful reasoning. In this case, the grasp planner is triggered to generate new grasps. 
\section{Experimental Results}
\label{sec:results}

\subsection{Dataset}
\label{sec:dataset}
Our part segmentation approach relies on depth data from an RGB-D camera to provide spatial context to the LLM. However, existing benchmarks such as PACO~\cite{paco2023} or Part-ImageNet~\cite{he2021partimagenet} only offer RGB images. Hence, for evaluation, we construct a custom dataset comprising 60 everyday household objects spanning 12 semantic classes, annotated with detailed part labels.
Our dataset includes five distinct instances per semantic class, varying in size and shape. Each object is captured in three different poses, resulting in a total of 180 RGB-D images. We manually annotate 2D ground truth masks for all relevant object parts. The selected classes with their parts are: bottle (cap, neck, body); hammer (handle, head); knife (handle, blade); mug (body, handle, rim); pan (handle, body); plier (handles, pivot, jaws); scissor (handles, pivot, blades); screwdriver (handle, shaft, tip); spoon (handle, bowl); spraying bottle (nozzle, trigger, body); stapler (base, upper arm); toothbrush (handle, brush head).
All data is collected in a clean, uncluttered tabletop setup, a common environment in robotic manipulation tasks. The dataset is publicly available to support further research.

\subsection{Implementation Details}
\label{sec:impl-details} 
Each LLM query follows a consistent structure that includes: (i) a task description (TD), (ii) supporting contextual and spatial information (SI), and (iii) a specification of the expected output format (OS). Although the general format is shared across modules, the supporting information is tailored to the requirements of each module.
To provide spatial context, the LLM is given 3D geometric information extracted via basic point cloud processing. This includes the object's center of mass, spatial bounds, and the orientation and length of its dominant axis. For detailed prompt examples, we refer the reader to our project website or supplementary material.\footnote{files named prompt.txt and example\_chat.txt}

In the reasoning stages of Part Segmentation, the structure of the LLM queries remains fixed, with only the task-specific problem description varying, for instance, labeling unlabeled clusters or detecting missing parts. The output is a string representing the part name for the query cluster. The supporting information for each query in this module includes the 3D geometric properties of the object and known parts, and the relevant query clusters or parts. Additional context, such as a tabletop setting, is provided as supporting information.
In the Grasp Selection module, the supporting information for the LLM query includes the 3D properties of the object and its parts, a list of candidate grasps, and the human interaction part provided by Task Reasoning. The output is the index of the best grasp.

\subsection{Part Segmentation Evaluation}
To assess the impact of our LLM-based reasoning algorithm on the performance of part segmentation, we consider common metrics from computer vision, namely intersection over the union (IoU) and F1-score. We also analyze the percentage of ground truth parts that are successfully detected, calling this metric the detection rate (DR).

 \begin{table}[ht]
 \vspace{-3pt}
 \centering
 \caption{\textsc{Part segmentation performance Comparison} between our LLM-enhanced method and the baseline (VLPart). Metrics are Detection Rate (DR), F1-score (F1), and Intersection over Union (IoU).}
 \resizebox{0.9\columnwidth}{!}{
 \begin{tabular}{l|rrr|rrr}
 \toprule
 \textbf{Object} & \multicolumn{3}{l|}{VLPart w. LLM-reasoning} & \multicolumn{3}{l}{VLPart} \\ \midrule
  & DR(\%) & F1 & IoU & DR(\%) & F1 & IoU \\
 \midrule
 Bottle & \textbf{66.68} & 0.76 & 65.96 & 61.67 & \textbf{0.77} & \textbf{66.82} \\
 Hammer & \textbf{100.00} & \textbf{0.88 }& \textbf{77.80} & 63.34 & 0.70 & 61.05 \\
 Knife & \textbf{96.66} & \textbf{0.92} & \textbf{84.30} & 90.00 & \textbf{0.92} & 84.02 \\
 Mug & \textbf{91.67 }& \textbf{0.78} & \textbf{65.83} & 87.50 & 0.76 & 65.22 \\
 Pan & \textbf{68.89} & \textbf{0.68} & \textbf{58.52} & 55.56 & 0.56 & 45.77 \\
 Plier & \textbf{88.23} & \textbf{0.80} &\textbf{ 67.65} & 70.59 & 0.73 & 58.86 \\
 Scissor & \textbf{71.11 }& \textbf{0.75} & \textbf{54.98} & 69.03 & 0.58 & 38.60 \\
 Screwdriver & \textbf{73.33} & \textbf{0.80 }&\textbf{ 65.89 }& 60.00 & 0.66 & 49.91 \\
 Spoon & \textbf{100.00 }& \textbf{0.90} & \textbf{81.09} & 91.66 & 0.88 & 79.36 \\
 Spraying bottle & \textbf{65.83} & \textbf{0.67} & \textbf{54.31} & 52.08 & 0.56 & 43.52 \\
 Stapler & \textbf{65.62} & \textbf{0.72} & \textbf{58.30} & 40.62 & 0.66 & 25.66 \\
 Toothbrush & 49.12 &\textbf{ 0.68} & \textbf{53.87} & \textbf{57.90} & 0.61 & 44.02 
 \\ \midrule
 \textbf{Mean} & \textbf{78.10} & \textbf{0.78 }& \textbf{77.82} & 66.66 & 0.70 & 69.99 \\
 \bottomrule
 \end{tabular}}
\label{tab:part-comparison}
\vspace{-10pt}
 \end{table}

For each object class, we present the mean values across all parts in Tab.~\ref{tab:part-comparison}. Notably, for some objects, such as the mug or bottle, VLPart already performs reasonably well, while for others, such as the hammer or plier, LLM-based reasoning leads to significant improvements across all metrics. Only the toothbrush shows a higher DR with VLPart. Although parts are detected more frequently, they’re often incorrect, as shown by the lower F1 and IoU scores. Furthermore, for, e.g. pan or hammer, one of the parts (the handle) is usually found. For these objects, the LLM-based reasoning ensures that at least two parts are found, i.e., for the hammer it detects the head, and for the pan its body. This effect is visible in the table through the significant increase in DR.

\begin{figure}[hb]
    \vspace{-18pt}
    \centering
\includegraphics[width=0.99\linewidth]{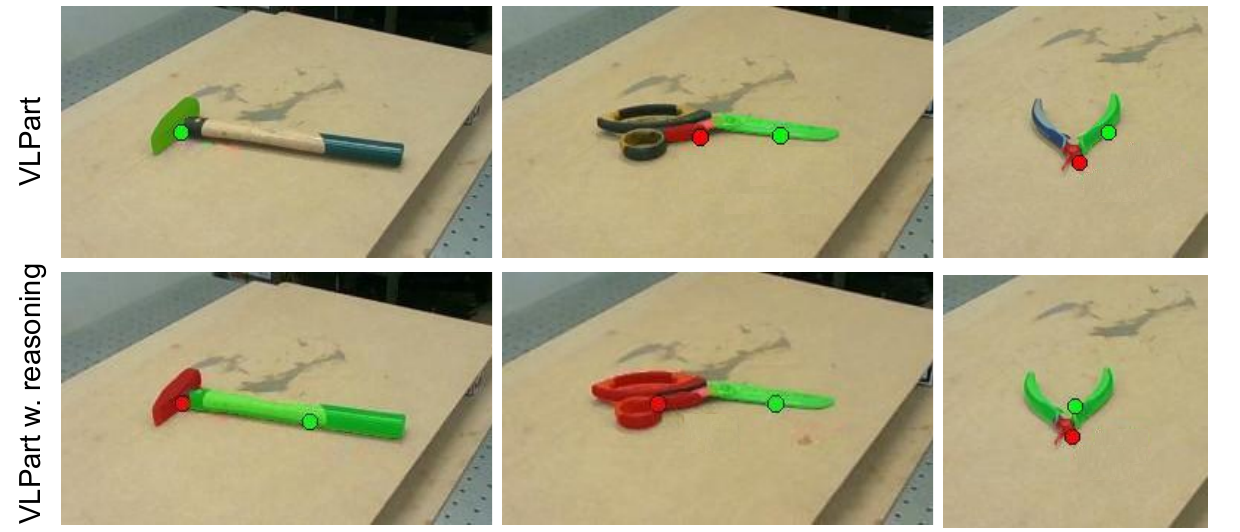}
    \vspace{-15pt}
    \caption{Qualitative part segmentation comparison. VLPart misses (handles for all) or mislabels parts (hammer head as handle).}
    \label{fig:llm-vl-part}
 %\vspace{-12pt}
\end{figure}

The results demonstrate that the LLM-based part segmentation reasoning improves the results across all evaluated metrics. Qualitative examples showing extreme cases are illustrated in Fig.~\ref{fig:llm-vl-part}. 
However, the effectiveness of the reasoning pipeline is limited by the quality of the initial segmentation.  If the initial segmentation is wrong and spatial reasoning cannot detect that (the screwdriver shaft being misidentified as the handle cannot be corrected geometrically), the LLM ends up reasoning with incorrect part assignments, leading to reversed labels for shaft and handle.

%%%%%%%%%%%%%%%%%%%%%%%%%%%%%%%
%%%%%%%%%%%%%%%%%%%%%%%%%%%%%%%
\subsection{Task-oriented handover evaluation}
\label{sec:toh-experiments}
Here, we describe the set of experiments we carried out to validate the LLM-Handover framework. First, we evaluate the reasoning capabilities of LLMs in two settings: high-level task understanding and grasp selection. Then, we carry out an ablation study for grasp selection to understand the impact of different types of information on the reasoning process. Furthermore, we compare our grasp selection results to existing baseline methods. Finally, we assess the performance of the framework on hardware and perform a user study.

For evaluation, we consider two categories of object and post-handover task pairs: conventional (e.g., use a hammer to hit a nail) and unconventional (e.g., use a screwdriver to play xylophone). We argue that conventional object-task pairs need a limited amount of understanding, while unconventional pairs require a thorough understanding of the object and task dynamics. Furthermore, we categorize our conventional object-task pairs into easy and complex. Easy pairs contain objects that have clearly defined parts for specific tasks and have parts that can easily be segmented by part segmentation. Complex pairs contain either objects with no clearly defined parts to grasp for the task, or have parts that cannot easily be detected by part segmentation. The list of the object-task pairs is given in Tab.~\ref{tab:obj-tasks}. 

\begin{table}[t]
\vspace{8pt}
\centering
\caption{\textsc{Object-task pairs} grouped by category.}
\begin{tabular}{l|cc|c}
\toprule
\scriptsize\textbf{Object/Task} & \multicolumn{2}{c}{\scriptsize\textbf{conventional}} & \scriptsize\textbf{unconventional} \\
\cmidrule(lr){2-3}
                       & \scriptsize \textit{easy}     & \scriptsize \textit{complex}&                  \\
\midrule
\scriptsize Hammer           & \scriptsize hammer       &                 &           \\
\scriptsize Knife            & \scriptsize cut          &                 &           \\
\scriptsize Mug              & \scriptsize drink        &                 &           \\
\scriptsize Screwdriver      & \scriptsize screw        &                 & \scriptsize hammer, play xylophone \\
\scriptsize Pan              & \scriptsize cook         &                 &           \\
\scriptsize Spoon            & \scriptsize stir         &                 & \scriptsize open lid of jar  \\ 
\scriptsize Scissor          &              & \scriptsize cut             &           \\
\scriptsize Plier            &              & \scriptsize pinch           &           \\
\scriptsize Stapler          &              & \scriptsize staple          &           \\
\scriptsize Bottle           &              & \scriptsize pour            &           \\
\scriptsize Spraying bottle  &              & \scriptsize spray           &           \\
\scriptsize Toothbrush       &              &\scriptsize  brush teeth     & \scriptsize push pin into a hole  \\
\bottomrule
\end{tabular}
\label{tab:obj-tasks}
\vspace{-20pt}
\end{table}

\subsubsection{High-level task understanding}
\label{sec:llm-reasoning}
In this section, we analyze the high-level reasoning skills for the handover and post-handover tasks. We evaluate the two most powerful OpenAI models, namely, GPT-4o (gpt-4o-2024-11-20) (G4o) and o1-preview (o1-p). During experiments, we notice that Task Reasoning performs better if the LLM is queried to also describe how the human will perform the post-handover task. This aligns with previous findings on chain-of-thought prompting for LLMs \cite{cot2022}, highlighting that correctly understanding the post-handover task plays an important role in the framework’s performance. Hence, in this analysis, we consider the following metrics: Post-handover task understanding accuracy (T), the percentage of correctly predicted parts on which the human (H) and robot (R) will grasp. T is computed by assessing the description of the post-handover task generated by the LLM. If the location of the human grasp and the part that makes contact with the environment are correct for the specific task, we consider that the task understanding of the LLM about the post-handover task is correct.
For each object-task pair, we run 10 experiments. 

As can be seen in Tab.~\ref{tab:task-reasoning}, there is a significant difference between conventional and unconventional tasks for all metrics. The understanding of tasks is mastered by both models for all conventional tasks. For the unconventional ones, GPT-4o has difficulties, while o1-preview understands most of them, except for the spoon to open a jar. We believe this is because the task requires significant knowledge of real-world dynamics to understand how one can use the spoon handle to achieve the task.
Furthermore, it is noticeable that even though the task is understood correctly, the models have difficulties predicting the correct grasping parts. This is especially the case for the robot grasp for complex conventional object-task pairs. We argue that this is because, despite the task being accurately described, it remains challenging to reason about the optimal robot grasp location for effective handover, especially when considering the constraints of the post-handover task. For example, in the case of a stapler to staple, the stapling task is described correctly. Nevertheless, GPT-4o predicts that the robot should grasp the middle of the base, which would make handover and direct use difficult. The o1-preview model correctly understands, in most cases, that the base should be grasped at the rear end, i.e., far from where the human pushes the top arm of the stapler. 
\begin{table}[t]
\vspace{8pt}
\setlength{\tabcolsep}{1.1pt}
\renewcommand{\arraystretch}{1.0}
\centering
\caption{\textsc{high-level task reasoning success rate} across different object-task pairs. Metrics are Task Understanding Accuracy (T), Robot (R), and Human (H) Grasp Prediction Accuracies.}
\begin{tabular}{lc|cc||cccc}
\toprule
\scriptsize \textbf{Model} & \scriptsize \textbf{Metric} & \multicolumn{2}{c||}{\scriptsize \textbf{conventional}}  & \multicolumn{4}{c}{ \scriptsize \textbf{unconventional}} \\
& &  &  & \scriptsize Spoon &\multicolumn{2}{c}{ \scriptsize Screwdriver}& \scriptsize Toothbrush  \\
\cmidrule(lr){6-7}
&  & \scriptsize easy& \scriptsize complex &\scriptsize \textit{open jar} 
& \scriptsize \textit{hammer} & \scriptsize \textit{play} & \scriptsize \textit{push pin} \\
\midrule

\multirow{3}{*}{G4o} 
& \scriptsize T & 100\% &100\% & 30\% & 80\% & 20\% & 50\% \\
& \scriptsize R &  93\% & 62\% & 0\% & 40\% & 10\% & 20\% \\
& \scriptsize H &  100\% & 95\% &  0\% & 10\% & 20\% & 30\% \\
\midrule

\multirow{3}{*}{o1-p}
& \scriptsize T & 100\% & 100\% & 20\% & 100\% & 90\% & 100\% \\
& \scriptsize R & 100\% & 87\% & 20\% &  90\% & 60\% &  70\% \\
& \scriptsize H &100\%   & 95\%  &  20\% & 100\% & 90\% & 100\% \\
\bottomrule
\end{tabular}
\vspace{-20pt}
\label{tab:task-reasoning}
\end{table}

\begin{table}[b!]
\vspace{-10pt}
\setlength{\tabcolsep}{2.0pt}
\renewcommand{\arraystretch}{1.0}
\centering
\caption{\textsc{UPPER: Comparison of robot grasp success rates} between LLM-Handover (ours) and baselines. \textsc{LOWER: Ablation study for Grasp Selection}. The metric reflects the rate of robot grasps not interfering with human interaction. (*) shows ground truth data is used.}
\begin{tabular}{l|cc||cccc||c}
\toprule
\scriptsize \textbf{Model}  & \multicolumn{2}{c}{\scriptsize \textbf{conventional}}  & \multicolumn{4}{c}{\scriptsize \textbf{unconventional}} \\
 &  &  & \scriptsize Spoon & \multicolumn{2}{c}{ \scriptsize Screwdriver} & \scriptsize Toothbrush & \scriptsize Average \\
\cmidrule(lr){5-6}
 & \scriptsize easy & \scriptsize \textit{complex} & \scriptsize \textit{open jar} & \scriptsize \textit{hammer} & \scriptsize \textit{play} & \scriptsize \textit{push pin} &  \\

\midrule
\scriptsize EdgeGrasp & 45\% & 44\% & 53\% & 80\% & 80\%&47\%& 49\% \\
\scriptsize GraspGPT & 49\% & 47\% & \textbf{58\%} & \textbf{87\%} & \textbf{87\%} & 40\% & 53\% \\
\scriptsize Heuristic & 87\% & 74\% & 21\% & 80\% & 80\% & 60\% & 74\% \\
\scriptsize \textbf{ours} & \textbf{97\%} & \textbf{82\%} & 21\% & \textbf{87\%} & \textbf{87\%} & \textbf{60\%} & \textbf{79\%} \\
%\multicolumn{18}{c}{\textbf{Comparison with variants of our method}} \\
\midrule
\midrule
\scriptsize LLM-nG*    
& 96\% &  93\%& \textbf{100\%}& \textbf{100\%} & \textbf{100\%} & \textbf{93\%}& 96\% \\
\scriptsize LLM-nH*    
& 96\% & 89\% & 53\% & 40\% & 60\% & 27\% & 80\% \\
\scriptsize \textbf{ours*}
& \textbf{99\%} & \textbf{98\%} & \textbf{100\%} & \textbf{100\%} & \textbf{100\%} & 87\% & \textbf{98\%} \\
\bottomrule
\end{tabular}
\label{tab:handover-comparison}
\end{table}

\subsubsection{Grasp selection reasoning}
\label{sec:llm-grasp}
In this section, we evaluate the Grasp Selection module (running gpt-4o-2024-11-20), in isolation, ensuring error-free inputs from upstream modules. Hence, Part Segmentation uses the hand-labeled masks from the dataset, while Task Reasoning is given the correct human and robot grasp parts manually. For Grasp Generation, we verify that at least one grasp lies on the correct part and retrigger grasp generation otherwise. We then manually label the grasp selected by LLM-Handover as good or bad. A grasp is considered good if it does not interfere with the human grasp required for the post-handover task. The success rate of correctly predicted robot grasps is used as a metric.

In total, 240 grasps are evaluated: one conventional task is considered for each of the 180 entries in the dataset, and an additional 60 grasps correspond to the 4 unconventional tasks. The results are shown in the last row of Tab.~\ref{tab:handover-comparison}. We achieve an overall robot grasp success rate of $98\%$. The grasp selection reasoning seems largely unaffected by the complexity of the object-task pairs. The only exception is the toothbrush to push a pin, with  $87\%$ success rate.
We attribute this to inaccuracies in the toothbrush point cloud, especially in thin regions, which result in slightly imprecise 3D information and subsequently impact performance. 

\subsubsection{Grasp selection ablation}
\label{sec:llm-grap-abl}
We ablate the effect of 3D geometric information and the human grasp area on the LLM's ability to select a correct grasp. We use the same setup as in Sec .~\ref{sec:llm-grasp} and consider the following cases: (i) LLM-Handover (ours): the LLM receives 3D geometric information, including the 3D centers of the found parts and object, and the desired human grasp part; (ii) LLM-nG: the LLM does not receive additional 3D geometric information, just the 3D centers of object and parts; (iii) LLM-nH: the LLM is not given the desired human grasp part.

The results are shown in the lower part of Tab.~\ref{tab:handover-comparison}. The ({$\star$}) indicates that ground truth information is provided as input. Providing the desired human grasp part to Grasp Selection noticeably improves the robot grasp accuracy, especially for unconventional tasks. For the conventional complex object-task pairs, the human grasp part leads to a gain of almost $10\%$. Geometric information yields a $5\%$ increase in the robot grasp accuracy for complex conventional object-tasks pairs, but has little effect on unconventional ones.
Overall, we notice that the human grasp part is the most informative cue for selecting a suitable grasp. 

\subsubsection{Comparison to state-of-the-art}
\label{sec:toh-sota}
We compare our framework against three baselines: GraspGPT \cite{tang2023graspgpt}, EdgeGrasp \cite{huang2022edge}, and a heuristic-based method. GraspGPT \cite{tang2023graspgpt} is a state-of-the-art task-oriented robot grasping framework shown to be effective for robot-human handovers. EdgeGrasp \cite{huang2022edge} is the grasp planner used in our framework. The heuristic-based method differs from LLM-Handover only in Grasp Selection: instead of using an LLM to select the most suitable grasp, grasps are reranked to avoid contact with the part the human intends to grasp, similar to \cite{contactHandover}.

We follow the setup and metric (correct robot grasp rate) from the previous sections, but Task Reasoning and Part Segmentation are performed by the framework, and Grasp Generation is not corrected to guarantee a valid grasp. As a result, the overall robot grasp accuracy of LLM-Handover drops from $98\%$ to $79\%$, as shown in Tab.~\ref{tab:handover-comparison}. This is most evident for spoon to open a jar and the toothbrush tasks. For the first, task reasoning, and for the latter, part segmentation is the limitation. 

\begin{figure}[t]
    \vspace{-4pt}
    \centering    \includegraphics[width=\linewidth]{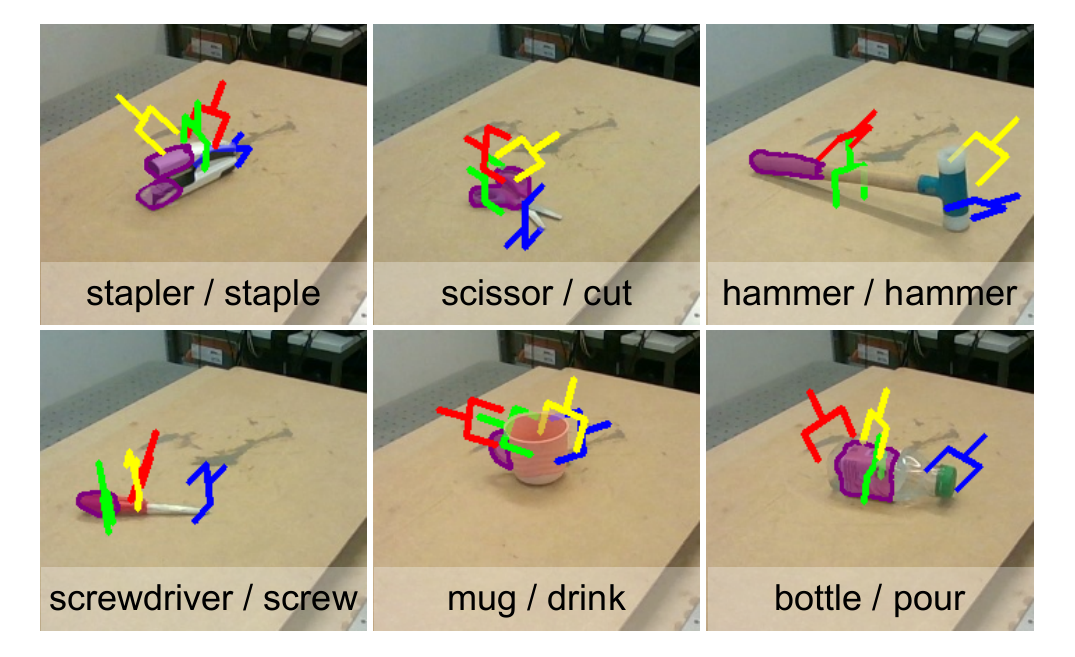}
    \vspace{-18pt}
    \caption{Qualitative comparison of selected robot grasps by method. Blue: LLM-Handover (ours); green: GraspGPT \cite{tang2023graspgpt}; red: EdgeGrasp \cite{huang2022edge}; yellow: heuristic. Purple highlights human grasp regions for the post-handover task, computed from video footage.}
    \label{fig:img_grasps}
\vspace{-15pt}
\end{figure}

The qualitative results from Fig.~\ref{fig:img_grasps} show that LLM-Handover consistently avoids the human grasp region, in contrast to the baselines. Quantitative results are summarized in the upper part of Tab.~\ref{tab:handover-comparison}. 
It can be seen that using knowledge about the post-handover task improves the overall robot grasp success rate. We achieve an overall success rate of $79\%$, while the closest baseline, the heuristic-based method, achieves $74\%$. While the latter is competitive, it lacks fine-grained spatial understanding. This is most visible if grasps are generated only on the human interaction part of an object; it randomly chooses a grasp, which is often close to the center, where the human would grasp. The Grasp Selection in LLM-Handover, on the other hand, infers that the grasp should be at the edges of the interaction part if that is the only part with feasible grasps.

Furthermore, except for spoon to open jar, LLM-Handover outperforms the baselines. In this particular case, LLM-Handover struggles with the high-level post-handover task understanding. The other two methods, which do not consider the post-handover task explicitly, achieve a success rate of around $50\%$, effectively equivalent to random grasp selection.
For the conventional tasks, the impact of post-handover task understanding is clearly visible. LLM-Handover achieves up to $97\%$ success rate.
For the unconventional screwdriver tasks, all methods have similar performance because grasps are generated mainly on the handle. Hence, even without reasoning, the probability of selecting a grasp on the handle is high. However, it is problematic for screwdriver to screw, where the shaft should be grasped by the robot, leading to poor results for EdgeGrasp \cite{huang2022edge} and GraspGPT \cite{tang2023graspgpt}.

%%%%%%%%%%%%%%%%%%%%%%%%%%%%%%%
%%%%%%%%%%%%%%%%%%%%%%%%%%%%%%%

\subsection{Hardware experiments}
\label{sec:hw-exp}
In this section, we integrate LLM-Handover into the existing robot motion planning framework and carry out hardware validation experiments and a comparative user study. 

\subsubsection{Experimental setup}
\label{sec:setup}
We validate our algorithm using a Franka Panda manipulator, with an external RealSense D-455 RGB-D camera monitoring the workspace. Motion planning is handled by MoveIt \cite{moveit2019}, ensuring collision avoidance with both the robot and its environment.

The human collaborator is asked to stay in a predefined area with respect to the robot. Since a real handover takes place, we have to compute a final handover pose. While we fix the handover position in a pre-defined area, the orientation of the object is computed by aligning the axis between the selected robot grasp and the predicted human grasp with the axis of the robot base pointing towards the human. This ensures that the part the human intends to grasp points towards the human. As we have a well-defined setup and know where the handover will take place, this holds. 

\subsubsection{Robot validation}
\label{sec:robot-exp}
For each of the 12 object classes from the dataset, we carry out 5 robot-human handovers for conventional post-handover tasks, totaling 60 experiments. Furthermore, we consider 3 unconventional tasks (screwdriver to hammer, spoon to open a jar, and toothbrush to push a pin into a hole), for which we also do 5 handovers, leading to a total of 75 handovers. We consider handover success rate (HSR) and the percentage of correctly predicted human (H) and robot grasps (R) as evaluation metrics.

As shown in Tab.~\ref{tab:hw-results}, we achieve an HSR of $83\%$ among all handovers, with an HSR of $60\%$ for unconventional tasks. The prediction accuracy for human grasps on the object is $91\%$, while robot grasps are correctly predicted in 
$88\%$ of the cases.
The difference in HSR between conventional easy and complex object-task pairs comes mainly from errors in part segmentation. The main reasons for failure are wrong part segmentation ($4\%$) or task reasoning ($4\%$), problems during motion planning ($5\%$), and limited generated grasp ($2\%$). 

We observe robustness to slight errors in part segmentation. In $11\%$ of the cases, most commonly for the stapler and toothbrush, Part Segmentation detected only a single part, typically the handle. Despite this limitation, failures occurred in only $40\%$ of those instances. For both objects, detecting only the handle often resulted in grasps on the sides of the object rather than the center. This had little impact on the stapler, but negatively affected the HSR for the toothbrush, because the human could not easily access the object when the robot grasped the end of the toothbrush handle.

\begin{table}[t]
\vspace{10pt}
\setlength{\tabcolsep}{2.5pt}
\renewcommand{\arraystretch}{1.2}
\centering
\caption{\textsc{Robot validation results} across various household object-task pairs.  Metrics are Handover Success Rate (HSR), Robot Grasp Prediction (R), and Human Grasp Prediction Accuracy (H).}
\resizebox{0.8\columnwidth}{!}{
\begin{tabular}{l|cc|ccc||c}
\toprule
\scriptsize \textbf{Metric} & \multicolumn{2}{c}{\scriptsize \textbf{conventional}}  & \multicolumn{3}{c}{ \scriptsize \textbf{unconventional}}  & \\
\midrule
 &  &  & \scriptsize Screwdriver & \scriptsize Spoon &\scriptsize Toothbrush &\scriptsize Average \\

 &  \scriptsize\textit{easy} &  \scriptsize \textit{complex}& \textit{hammer} & \textit{open jar} & \textit{push pin}&   \\

\midrule
\scriptsize HSR & 93\% & 77\% & 80\% & 40\% & 60\% & 83\%  \\
\scriptsize H & 97\% & 93\% & 80\% & 40\% & 80\%& 91\%  \\
\scriptsize R & 97\% & 90\% & 80\% & 40\% & 60\% & 88\% \\
\bottomrule
\end{tabular}
}
\label{tab:hw-results}
\vspace{-20pt}
\end{table}

\subsubsection{User study}
\label{sec:uesr-study}
To validate our findings and obtain subjective feedback, we carried out a user study with 14 participants, reviewed and approved by the ETH Zürich ethics committee under application \textit{25 ETHICS-021}. Each participant carried out seven runs of two robot-human handovers. One run used a baseline (GraspGPT \cite{tang2023graspgpt}), while the other used LLM-Handover. Seven objects and their corresponding post-handover tasks were predefined. Of the seven tasks, one was unconventional: screwdriver to hammer.

The order of the objects was randomized and each participant performed a handover and completed the predefined task for each object. The method order (LLM-Handover vs. baseline) was fixed per participant but randomly assigned across participants to allow for overall preference comparison at the end. Participants were unaware of which method was used first or second. After each handover, they noted if regrasps were necessary to complete the predefined task comfortably. With this, the regrasp rate (R) was computed as the percentage of handovers needing adjustment before the post-handover task. Additionally, they provided binary ratings on: run preference (Pf), perceived task understanding by the robot (U), and object presentation quality (P), e.g., easily accessible, oriented properly.

To compute the object orientation at the handover location, the predicted human grasp is necessary. However, the baseline only predicts the best robot grasp for handover and not the human grasp. To overcome this, we queried GraspGPT \cite{tang2023graspgpt} a second time with the post-handover task of the human as the query task to get an appropriate grasp location for the human. We used this grasp location as the predicted human grasp location and followed the heuristic described in Sec.~\ref{sec:setup} to compute the orientation. The participants were shown the predefined handover location to know where to expect the object. We prepared a mockup board to allow the participants to do the tasks as realistically as possible. 
%(as shown in Fig.~\ref{fig:handover}). 
\begin{table}[!b]
\vspace{-8pt}
\centering
\caption{\textsc{User study comparison} between GraspGPT and LLM-Handover. using object presentation quality (P), participant preference (Pf), regrasps rate (R), and perceived task understanding (U) as metrics.}
\resizebox{0.9\columnwidth}{!}{\begin{tabular}{l|rrrr|rrrr}
\toprule
\scriptsize \textbf{Method} & \multicolumn{4}{c|}{\scriptsize \textbf{GraspGPT}} & \multicolumn{4}{c}
{\scriptsize \textbf{LLM-Handover}} \\
\cmidrule{2-5}
\cmidrule{6-9}
\scriptsize \textbf{Metrics (\%)} &\scriptsize P & \scriptsize Pf & \scriptsize R$\downarrow$ & \scriptsize U & \scriptsize P & \scriptsize Pf & \scriptsize R$\downarrow$ & \scriptsize U \\
\midrule
\scriptsize Bottle \textit{(pour)}& \textbf{64} & 33 & 36 & 47 & 36 & \textbf{67} & \textbf{29} & \textbf{53} \\
\scriptsize Mug \textit{(drink)}& 50 & 43 & \textbf{7} & \textbf{53} & \textbf{79} & \textbf{57} & 14 & 47 \\
\scriptsize Spoon \textit{(stir)}& 14 & 24 & 71 & 40 & \textbf{64} & \textbf{76} & \textbf{21} & \textbf{60} \\
\scriptsize Pan \textit{(cook)}& 43 & 33 & 36 & 37 & \textbf{64} & \textbf{67} & \textbf{21} & \textbf{63} \\
\scriptsize Hammer \textit{(hammer)}& 36 & 41 & \textbf{14} & 44 & \textbf{64} & \textbf{59} & \textbf{14} & \textbf{56} \\
\scriptsize Screwdriver \textit{(screw)} & 21 & 12 & 36 & 18 & \textbf{86} & \textbf{88} & \textbf{0} & \textbf{82} \\
\scriptsize Screwdriver \textit{(hammer)} & 36 & 29 & 29 & 41 & \textbf{43} & \textbf{71} & \textbf{21} & \textbf{59} \\
\midrule
\scriptsize Average ($\%$) & 38 & 31 & 36 & 40 & \textbf{62} & \textbf{69 }& \textbf{14} & \textbf{60}\\
\bottomrule
\end{tabular}}
\label{tab:user-study}
\end{table}
As indicated in Tab.~\ref{tab:user-study}, LLM-Handover outperforms the baseline in almost all cases. Hence, understanding the task the human intends to perform after the handover improves the subjective preference for the interaction. Solely for the mug, the baseline performs better in the number of regrasps and understanding. We attribute this to the mug being presented in an unusual pose (handle facing the human, but with the opening upside-down), which negatively impacted the results for understanding and required regrasping. This issue arises because the current grasp selection does not account for orientation relative to task affordances, and the handover heuristics used for orientation computation lack constraints on mug alignment. Despite this, $85.71\%$ of participants preferred our method overall over the baseline ($14.29\%$). 

\section{Limitations}
\label{sec:Discussions}
Even though the experiments confirm that the framework generalizes to different object-task pairs and has a solid understanding of the post-handover task, especially for the conventional tasks, there are still some cases in which the method does not perform well. LLM reasoning struggles with the unconventional object-task pairs (e.g., using a spoon to open a jar). These scenarios require a nuanced understanding of physical dynamics and affordances, which current models may not reliably capture. 
Another limitation of our method is part segmentation. Currently, only big/significant object parts are segmented. However, sometimes sub-part accuracy is needed to grasp correctly.%, e.g., the bottom of the bottle body or the rear part of the stapler base. 
While the LLM can reason about this and often correctly deliver sub-part accuracy for grasping, the part segmentation algorithm remains a bottleneck.

From the user study, we observed that the pre-defined handover location is a limitation, as people have different handover location preferences for different objects. Moreover, our current setup does not account for human hand motion during the handover. Participants noted this felt unnatural, especially when they tried to reach for the object and the robot ignored their motion. A reactive system that adapts in real-time to human or hand pose would improve the naturalness of the interaction.
Additionally, for the stirring or screwing task, the appropriate handover orientation and the subsequent human grasp depend on the amount of force needed. Hence, more information about the task to be done could improve the interaction quality. 
Furthermore, computations currently take $3-25~s$, depending on how much part segmentation reasoning is necessary and how many grasps are returned by Grasp Generation. Although this did not bother study participants, it is impractical for real-world deployment.

\section{Conclusions and future work}
\label{sec:Conclusions}

In this work, we introduce LLM-Handover, a novel task-oriented robot-human handover framework that uses LLM-based reasoning and part segmentation to select the most suitable grasp for handover. Using a custom RGB-D dataset of 12 household object classes, we demonstrate that spatial-aware reasoning improves part segmentation performance, especially for challenging object parts missed by standard methods like VLPart \cite{peize2023vlpart}.
Beyond perception, our framework enables high-level task understanding and informed grasp selection. Through ablation studies and comparisons to baselines such as GraspGPT \cite{tang2023graspgpt} and EdgeGrasp \cite{huang2022edge}, we demonstrate that reasoning improves grasp success. 

Moreover, we show the effectiveness of our framework through hardware experiments and a user study comparing it to GraspGPT \cite{tang2023graspgpt}. In hardware experiments, our method achieves an $83\%$ handover success rate in a zero-shot setting, without any task-specific training.
The user study confirms the benefits of our approach, with $86\%$ of users preferring our system over the baseline. These results highlight the potential of LLM-driven reasoning to improve both the intelligence and intuitiveness of human-robot collaboration.

In future work, we aim to address current limitations and incorporate more task-specific constraints, such as keeping a cup upright to avoid spilling or maintaining tool orientation for immediate use, to enable more effective handovers.

\addtolength{\textheight}{0cm}   % This command serves to balance the column lengths
                                  % sure that you do not shorten the text height too much.

\bibliographystyle{IEEEtran}
\bibliography{mybibfile}

\end{document}